\documentclass[11pt]{article}
\usepackage{eacl2017}
\usepackage{times}
\usepackage{url}
\usepackage{latexsym}
\usepackage{graphicx}
\usepackage{tabularx,ragged2e,caption}
\usepackage{multicol}

\eaclfinalcopy % Uncomment this line for the final submission
\def\eaclpaperid{412} %  Enter the acl Paper ID here

\title{Automatically augmenting an emotion dataset improves classification using audio}

\author{Egor Lakomkin \hspace{0.5cm} Cornelius Weber\hspace{0.5cm} Stefan Wermter \\
  Department of Informatics, Knowledge Technology Group\\ University of Hamburg \\ Vogt-Koelln Str. 30, 22527 Hamburg, Germany \\
  {\tt \{lakomkin, weber, wermter\}@informatik.uni-hamburg.de}
  }

\date{}

\begin{document}
\maketitle
\begin{abstract}
  In this work, we tackle a problem of speech emotion classification. One of the issues in the area of affective computation is that the amount of annotated data is very limited. On the other hand, the number of ways that the same emotion can be expressed verbally is enormous due to variability between speakers. This is one of the factors that limits performance and generalization. We propose a simple method that extracts audio samples from movies using textual sentiment analysis. As a result, it is possible to automatically construct a larger dataset of audio samples with positive, negative emotional and neutral speech. We show that pretraining recurrent neural network on such a dataset yields better results on the challenging EmotiW corpus. This experiment shows a potential benefit of combining textual sentiment analysis with vocal information.
\end{abstract}

\section{Introduction}

Emotion recognition recently gained a lot of attention in the literature. The evaluation of the human emotional state and its dynamics can be very useful for many areas such as safe human-robot interaction and health care. While recently deep neural networks achieved significant performance breakthroughs on tasks such as image classification \cite{Simonyan2014VeryRecognition}, speech recognition \cite{HannunDeepRecognition} and natural language understanding \cite{Sutskever2014SequenceNetworks}, the performance on emotion recognition benchmarks is still low. 
A limited amount of annotated emotional samples is one of the factors that negatively impacts the performance. While obtaining such data is a cumbersome and expensive process, there are plenty of unlabelled audio samples that could be useful in the classifier learning \cite{ghosh2015learning}. \par
    The majority of recent works use neural networks combining facial expressions and auditory signals for emotion classification \cite{barros2015emotional,yao2015capturing,chao2016audio}. There is a clear benefit of merging visual and auditory modalities, but only in those situations when the speaker's face can be observed.  In \cite{Hines2015Twitter:Recognition} it was shown that incorporating linguistic information along with acoustic representations can improve performance. Semantic representations of spoken text can help in emotional class disambiguation, but in this case, the model will rely on the accuracy of the speech-to-text recognition system. Pretraining convolutional neural network \cite{ebrahimi2015recurrent} on an external dataset of faces improves the performance of the emotion classification model. However, the problem of augmenting emotional datasets with audio samples to improve the performance of solely audio processing models remained unsolved.\par 
    Our motivation for this paper was to fill this gap and conduct experiments on automatically generating a larger and potentially richer dataset of emotional audio samples to make the classification model more robust and accurate. In this work, we describe a method of emotional corpus augmentation by extracting audio samples from the movies using sentiment analysis over subtitles. Our intuition is that there is a significant correlation between the sentiment of spoken text and an actually expressed emotion by the person. Following this intuition we collect positive, neutral and negative audio samples and test the hypothesis that such an additional dataset can be useful in learning more accurate classifiers for the emotional state prediction. Our contribution is two-fold: a) we introduce a simple method to extract automatically positive and negative audio training samples from full-length movies b) we demonstrate that using an augmented dataset improves the results of the emotion classification.

\section{Models and experimental setup}

\subsection{Dataset}
For our experiment, we have used the EmotiW 2015 dataset \cite{dhall2015video}, which is a well-known corpus for emotional speech recognition composed of short video clips annotated with categorical labels such as \textit{Happy, Sad, Angry, Fear, Neutral, Disgust and Surprise}.  Each utterance is approximately 1-5 seconds in duration. The EmotiW dataset is considered as one of the most challenging datasets as it contains samples from very different actors and the lighting conditions, background noise and other overlapping sounds make the task even more difficult. The training set and the validation set contains 580 and 383 video clips respectively. We have used the official EmotiW validation set to report the performance as the test set labels were not released and 10\% of the official training set as validation set for neural network early stopping.

\subsection{Generating emotional audio samples}

As a source for emotional speech utterance candidates, we use full-length movies taking the list of titles from the EmotiW corpus. For each of the films, there are subtitles available, which can be treated as a good approximation of a spoken text, even though sometimes there can be inaccuracies as producing subtitles is a manual process. Our intuition is that the movies contain a large variety of auditory emotional expressions by many different speakers and is a potentially valuable source of emotional speech utterances. For each of the movies, sentiment score was calculated for each of the subtitle phrases at the time of the utterance with the NLTK \cite{BirdKleinLoper09} toolkit. Sentiment score represents how positive or negative the text segment is. The NLTK sentiment analyzer was used for simplicity and effectiveness. Phrases longer than 100 characters and shorter than four words were filtered out to avoid having very long or very short utterances. Subtitle phrases with polarity score higher than 0.7 were treated as positive samples and the ones with sentiment score lower than -0.6 as negative samples. The thresholds were selected empirically to make the number of the positive and negative samples balanced. As the majority of the phrases were assigned a value of sentiment close to 0, we treated them as neutral and used only a random subsample of it. Corresponding audio samples were cut from the movie with respect to the timings of the subtitle phrase. Overall 2100 positive, negative and neutral speech utterances were automatically selected from 59 movies and used as the additional dataset for emotion classification for binary tasks and as a dataset for model pre-training in multi-class setup.

\begin{figure}
  \centering
  \includegraphics[scale=0.47]{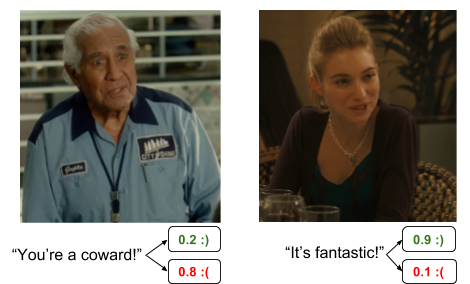}
  \caption{Vizualization of the process of extraction of positive and negative speech utterances, based on sentiment analysis of the subtitles.}
\end{figure}

\subsection{Features extracted}
We extracted FFT (Fast Fourier Transform) spectrograms from the utterances with a window length of 1024 points and 512 points overlap. Frequencies above 8kHz and below 60Hz were discarded as higher frequencies usually contain more noise and a log-scale in the frequency domain was used as emphasizing lower frequencies appears to be more significant for the emotional state prediction  \cite{ghosh2015learning}. Maximum length of the utterance in the dataset is 515 frames.\par
\subsection{GRU model}
The Gated-recurrent unit (GRU) \cite{bahdanau2014neural} is a recurrent neural network \cite{elman1991distributed} model trained to classify a sequence of input vectors. One of the main reasons for its success is that the GRU is less sensitive to the vanishing gradient problem during training, which is especially crucial for acoustic processing as the length of the sequences can easily reach hundreds or even thousands of time steps, as opposed to NLP tasks. \par
As a first stage, a single layer bi-directional GRU model has been used in our experiments with a 32 dimension cell size. Temporal mean pooling over all intermediate hidden memory representations was used to construct the final memory vector.
\begin{equation}
z = \sigma(x_tU^z + s_{t-1}W^z)
\end{equation}
\begin{equation}
r = \sigma(x_tU^r + s_{t-1}W^r)
\end{equation}
\begin{equation}
h^{fw} = tanh(x_tU^h+(s_{t-1}^{fw}\circ r)W^h)
\end{equation}
\begin{equation}
h^{bw} = tanh(x_tU^h+(s_{t+1}^{bw}\circ r)W^h)
\end{equation}
\begin{equation}
s_t^{fw} = (1-z)\circ h_t^{fw} + z \circ s_{t-1}^{fw}
\end{equation}
\begin{equation}
s_t^{bw} = (1-z)\circ h_t^{bw} + z \circ s_{t+1}^{bw}
\end{equation}
\begin{equation}
s_t = concat([s_t^{fw},s_t^{bw}])
\end{equation}
\begin{equation}
c = \frac{\sum_{t=1}^{T}s_t}{T}
\end{equation}
In these equations, the $c$ vector is used to represent the  whole speech utterance as an average of intermediate memory vectors $s_t^{fw}$ and $s_t^{bw}$, where $fw$ index corresponds to forward GRU execution and $bw$ for backward. $x_t$ is a spectrogram frame, $r$ and $z$ represent resent and update gates and $s_t$ is a GRU memory representation at timestamp $t$, following notation in \cite{bahdanau2014neural}. We have used Keras \cite{chollet2015keras} and Theano \cite{Bastien-Theano-2012} frameworks for our implementation.

\begin{figure}[h!]
  
  \centering
  \includegraphics[scale=0.28]{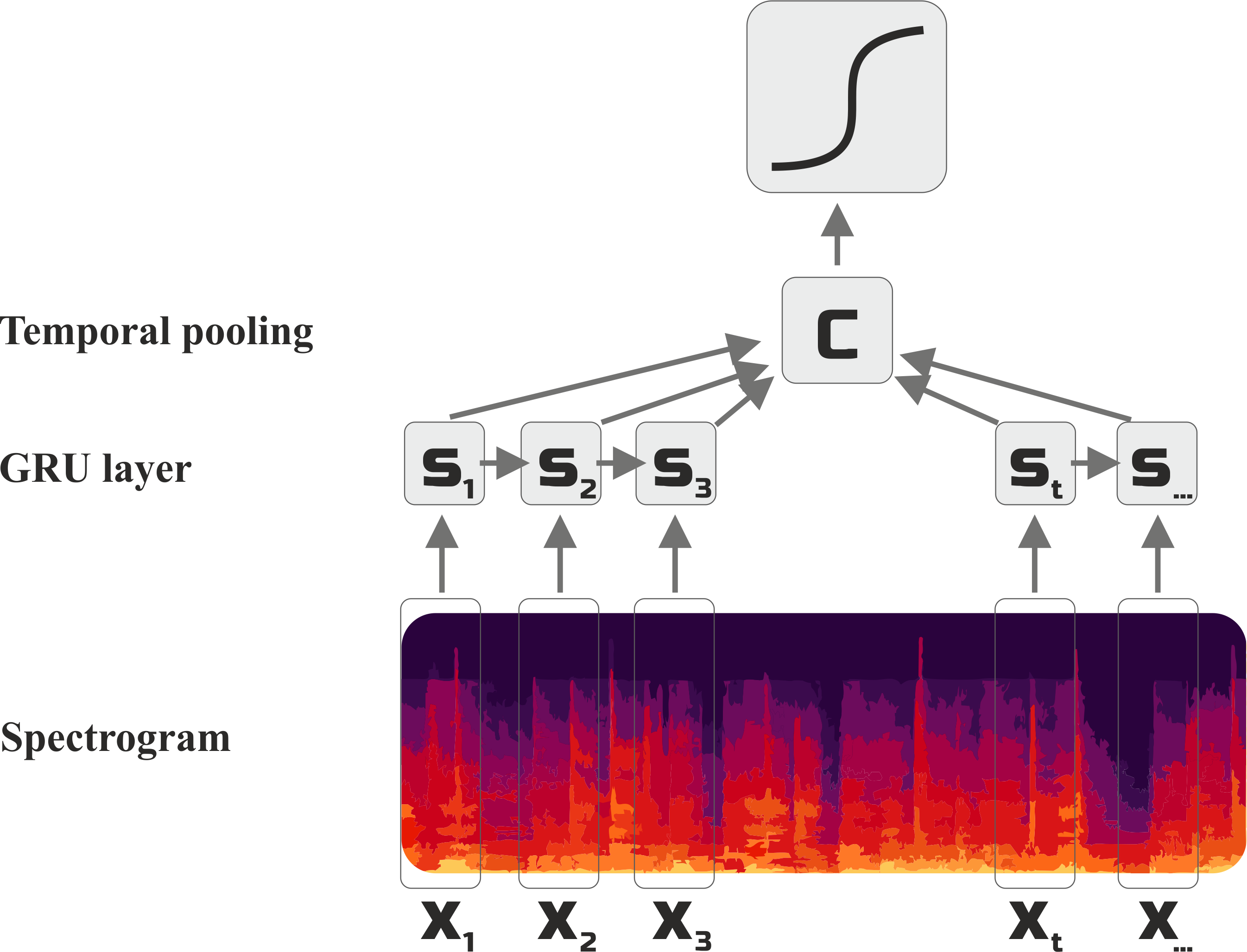}
  \caption{Recurrent neural network model for emotion speech utterance classification with temporal pooling.}
\end{figure}

\subsection{Transfer learning}
In multi-class setup, firstly we trained the neural network on the augmented corpus predicting labels generated by sentiment analyzer. We refer to it as a pre-trained network. As our goal is to predict emotional categories like happy or anger, we afterward replaced the softmax layer of the pre-trained network comprised of positive, negative and neutral classes with the new softmax layer for emotion prediction with angry, happy, sad and neutral classes. By using such a procedure, GRU layer hopefully can grasp meaningful representation of positive, neutral and negative speech which, as a result, will be helpful for emotion classification by means of transfer learning. Fine-tuning was done on the training data of the EmotiW corpus.

\subsection{Results}
We compare our results in three binary emotion classification tasks: \textit{happy-vs-fear}, \textit{happy-vs-disgust} and \textit{happy-vs-anger} and multi-class setup, where we considered \textit{Happy}, \textit{Angry}, \textit{Sad} and \textit{Neutral} samples. For each of the tasks we treated generated negative samples as either \textit{fear}, \textit{disgust} or \textit{anger} samples respectively and positive samples as \textit{happy}. For the multi-class setup, we follow the transfer learning routine by adapting neural network trained on the augmented data to the 4-way emotional classification. Accuracy is reported for binary tasks and F-score for multi-class setup. Results are presented in Table 1. By using automatically generated emotional samples there is a slight decrease in the accuracy for \textit{happy-vs-anger} task and an improvement in the accuracy for \textit{happy-vs-fear} and \textit{happy-vs-disgust} tasks. Also, in our experiments, temporal pooling worked significantly better than using the memory vector at the last time step.

\begin{table}[!htbp]
\centering
\caption{Utterance level emotion classification performance (accuracy) in 3 binary tasks: happy vs fear, happy vs angry and happy vs disgust. Also, multi-class performance (F-measure) is reported with 4 basic emotions: Angry, Happy, Sad and Neutral. BM - baseline method without augmentation, PM - proposed method with augmentation.}
\bigskip
\begin{tabular}{l*{2}{c}}

\textbf{Experiment}              & \textbf{BM} & \textbf{PM} \\
\hline
Binary classification: \\

Happy vs Fear & 58.7 & \textbf{66.1 } \\
\hline
Happy vs Angry        &     \textbf{70.7}  & 68.9  \\
\hline
Happy vs Disgust        &     61.1  & \textbf{64.1 } \\
\hline
Multi-class: \\
Angry, Happy, Sad, Neutral        &     36  & \textbf{38 } \\
\hline

\end{tabular}

\end{table}

\section{Conclusion}

In this paper, we proposed a novel method for automatically generating positive, neutral and negative audio samples for emotion classification from full-length movies. We experimented with three different binary classification problems: happy vs anger, happy vs fear and happy vs disgust and found that for the latter two there is an improvement in the accuracy on the official EmotiW validation set. Also, we observed the improvements of the results in multi-class setup. We found that the augmented larger dataset even though contains noisy and weak labels, contribute positively to the accuracy of the classifier.\par
For future work, we want to explore jointly learning sentiment and acoustic representations of the spoken text, which appears to be beneficial for accurate speech emotion classification, as it allows to deal with the ambiguity of the spoken text sentiment.

\section*{Acknowledgments}

\bibliography{Citations,Mendeley}
\bibliographystyle{eacl2017}

\end{document}